\documentclass[11pt]{article}
\usepackage[utf8]{inputenc}
\usepackage[final]{acl}
\usepackage{multirow, makecell}
\usepackage{amsmath}
\usepackage{algorithm}
\usepackage{algpseudocode}
\usepackage{amssymb}
\usepackage{enumitem}

\usepackage{times}
\usepackage{latexsym}
\usepackage{graphicx}

\usepackage{xcolor,colortbl}
\usepackage[T1]{fontenc}
\usepackage[utf8]{inputenc}
\usepackage{tcolorbox}
\tcbuselibrary{breakable}
\usepackage{mwe}
\usepackage{tabu}

\usepackage{microtype}

\definecolor{light-gray}{gray}{0.80}
\definecolor{custom-yellow}{RGB}{255,255,102}
\definecolor{light-yellow}{RGB}{255,255,150}
\definecolor{lightest-yellow}{RGB}{255,255,204}
\definecolor{light-green}{RGB}{153,255,153}
\definecolor{light-red}{RGB}{255,153,153}

\newcommand{\thor}{\textsc{THoR}}
\newcommand{\thorState}{\thor{}\textsc{\textsubscript{state}}}
\newcommand{\thorCause}{\thor{}\textsc{\textsubscript{cause}}}
\newcommand{\thorCauseRR}{\thor{}\textsc{\textsubscript{cause-rr}}}

\newcommand{\trainData}{\textsc{train}\textsubscript{json}}
\newcommand{\testData}{\textsc{test}\textsubscript{json}}
\newcommand{\labelsMath}{E}
\newcommand{\labelsAllMath}{E'}
\newcommand{\dState}{$D\text{\textsubscript{state}}$}
\newcommand{\dCause}{$D\text{\textsubscript{cause}}$}
\newcommand{\dTuple}{t}
\newcommand{\dLabels}{L}        
\newcommand{\fdev}{$F1({\labelsAllMath{}})$}        

\title{nicolay-r at SemEval-2024 Task 3: 
Using Flan-T5 for
Reasoning Emotion Cause in Conversations 
with Chain-of-Thought on Emotion States}
\author{
  Nicolay Rusnachenko \\
  Newcastle Upon Tyne, England \\
  {\texttt{rusnicolay@gmail.com}}
  \\\And
  Huizhi Liang  \\
  Newcastle University  \\
  Newcastle Upon Tyne, England \\
  \texttt{huizhi.liang@ncl.ac.uk} \\
  }

\date{January 2024}

\begin{document}

\maketitle

\begin{abstract}
Emotion expression is one of the essential traits of conversations. 
It may be self-related or caused by another speaker.
The variety of reasons may serve as a source of the further emotion causes:
conversation history, 
speaker's emotional state,
etc.
Inspired by the most recent advances in 
Chain-of-Thought,
in this work, we exploit
the existing three-hop reasoning approach
(\thor{})
to perform large language model instruction-tuning 
for answering:
emotion states (\thorState{}), and
emotion caused by one speaker to the other (\thorCause{}).
We equip
\thorCause{}
with the reasoning revision (\textsc{rr}) for devising a reasoning path
in fine-tuning. 
In particular, we rely on the 
annotated 
speaker emotion states to revise reasoning path. 
Our final submission, based on
Flan-T5\textsubscript{base} (250M)
and the rule-based span correction technique,
preliminary tuned with 
\thorState{} and
fine-tuned with
\thorCauseRR{}
on competition training data,
results in $3\textsuperscript{rd}$ and $4\textsuperscript{th}$ places
($F1_\text{proportional}$) and 
$5\textsuperscript{th}$ 
place ($F1_\text{strict}$) 
among 15 participating teams.
Our
\thor{} 
 implementation fork is publicly available:~\url{https://github.com/nicolay-r/THOR-ECAC}

\end{abstract}

\section{Task Overview}

Extracting potential causes that lead to emotion expressions in text is the crucial aim of Emotion Cause Extraction (ECE) domain~\cite{xia2019emotion}.
In particular, the
SemEval-2024~Task~3~\cite{wang-EtAl:2024:SemEval20244} is aimed at
emotion-cause pair analysis in conversations from 
the sitcom \textit{Friends}. 
The conversations are organized into Emotion-Cause-in-Friends dataset~\cite{wang2023multimodal} and includes the JSON-formatted 
training (\trainData{}) and evaluation (\testData{}) parts.
The authors propose $6$ emotion classes to annotate: 
(i)~speaker emotion states, and 
(ii)~emotion caused by one utterance to the other.
These classes are: $\labelsMath{}=\{\text{\textsc{\small surprise, sadness, joy, disgust, fear, anger}}\}$, and {\small \textsc{neutral}} for absence of emotion.
We denote $\labelsAllMath{}=\labelsMath{}~\cup~\{\text{\small \textsc{neutral}}\}$ as a complete set.

Among the several subtasks of ECAC-2024, in this paper we focused on 
\textit{Subtask~1}: textual emotion-cause pair extraction in conversations.
In this subtask,
each conversation represents a list of utterances.
Every utterance ($u$) yields the following:
utterance text ($u_{text}$), 
speaker name ($u_{speaker}$), 
emotion state ($u_{state}\in\labelsAllMath{}$), 
and \textsc{id} ($u_{id}$).
The annotation of the emotion cause pairs represents a list $P=[ p_1 \ldots p_{|P|}]$, 
in which each pair $p\in P$
is a labeled source-target\footnote{Spans-prediction is beyond the scope of our methodology.} 
tuple
$p=\left<u^{src}, u^{tgt}, e_{c}\right>$,
where
$e_{c} \in \labelsMath{}$.

We initiate our studies by analyzing the training data (\trainData{}) for the subject of annotated emotion-cause pairs $\left<u^{src}, u^{tgt}\right>$ in it, and report:
\begin{enumerate}
    \item Quantitative statistics 
    of the mentioned emotion-cause pairs (Table~\ref{tab:train_analysis_s});
    \item Distance statistics (in utterances) between 
    $u^{src}$ and $u^{tgt}$ (Table~\ref{tab:train_analysis_dist});
    \item Distribution statistics between speaker state ($u_{state}$) and emotion \textit{speaker causes} ($e^{u\to*}$)
    (Table~\ref{tab:train_analysis_se}).
\end{enumerate}

\begin{table}[!t]
    \centering
    \small
    \setlength{\tabcolsep}{4pt} 
   \begin{tabular}{p{6.1cm}|r}
    Parameter           & Value \\
    \hline 
    Conversations (total)                                        & 1374 \\
    Emotion causes pairs per conversation                        & 6.46 \\
    Emotion causes pairs in annotation (total)                   & 8879\\
    \hspace{2mm} Self-cause per conversation (\% from total)     & \textbf{51.86\%} \\
    \hspace{2mm} Self-cause by different utterance (\% from total) & 12.83\% \\
    \hline
    \end{tabular} 
    \caption{Quantitative statistics of the emotion-cause pairs 
    in the competition training data (\textsc{train}\textsubscript{json})}
    \label{tab:train_analysis_s}

    \bigskip
    
    \begin{tabular}{l|r|rrrrr}
    Parameter        & future    & \multicolumn{5}{c}{past}   \\
    \hline
    $\delta = u^{tgt}_{id}-u^{src}_{id}$  & $<$ 0 & 0    & 1    & 2    & 3             & 4    \\
    \hline 
    Causes count     &  377 & 4605 & 2759 & 810  & 332           & 160  \\
    \hspace{2mm}Average per $\delta$& 0.12 & 3.35 & 2.01 & 0.59 & 0.24          & 0.12 \\
    \hspace{2mm}Covering (\%)    &   -- & 51.9 & 82.9 & 92.1 & \textbf{95.8} & 97.6 \\
    \hline
    \end{tabular}
    \caption{Distance statistics ($\delta$) (in utterances) between source ($u^{src}$) and target ($u^{tgt}$) of emotion-cause pairs in the competition training data (\textsc{train}\textsubscript{json})}
    \label{tab:train_analysis_dist}
    
    \bigskip
    
    \begin{tabular}{l|cccccc}
    $u_{state} \backslash e^{u\to*}$    & {\color[HTML]{000000} \textsc{joy}}                         & {\color[HTML]{000000} \textsc{sur}}                         & {\color[HTML]{000000} \textsc{ang}}                         & {\color[HTML]{000000} \textsc{sad}}                         & {\color[HTML]{000000} \textsc{dis}}                         & {\color[HTML]{000000} \textsc{fea}}                         \\
    \hline
    total             & 2653	& 2092	& 1984	& 1336	& 518	& 296 \\
    \hline
    
    {\color[HTML]{000000} \textsc{joy}} & \cellcolor[HTML]{5BC5CD}{\color[HTML]{000000} .89} & \cellcolor[HTML]{F4FCFC}{\color[HTML]{000000} .06} & \cellcolor[HTML]{FAFEFE}{\color[HTML]{000000} .03} & \cellcolor[HTML]{FEFFFF}{\color[HTML]{000000} .01} & \cellcolor[HTML]{FEFFFF}{\color[HTML]{000000} .01} & \cellcolor[HTML]{FFFFFF}{\color[HTML]{000000} .00} \\
    {\color[HTML]{000000} \textsc{surprise}} & \cellcolor[HTML]{F3FBFC}{\color[HTML]{000000} .07} & \cellcolor[HTML]{6FCCD3}{\color[HTML]{000000} .78} & \cellcolor[HTML]{F3FBFC}{\color[HTML]{000000} .07} & \cellcolor[HTML]{FAFEFE}{\color[HTML]{000000} .03} & \cellcolor[HTML]{FAFEFE}{\color[HTML]{000000} .03} & \cellcolor[HTML]{FCFEFE}{\color[HTML]{000000} .02} \\
    {\color[HTML]{000000} \textsc{anger}} & \cellcolor[HTML]{FEFFFF}{\color[HTML]{000000} .01} & \cellcolor[HTML]{F3FBFC}{\color[HTML]{000000} .07} & \cellcolor[HTML]{66C9D0}{\color[HTML]{000000} .83} & \cellcolor[HTML]{F4FCFC}{\color[HTML]{000000} .06} & \cellcolor[HTML]{FCFEFE}{\color[HTML]{000000} .02} & \cellcolor[HTML]{FCFEFE}{\color[HTML]{000000} .02} \\
    {\color[HTML]{000000} \textsc{sadness}} & \cellcolor[HTML]{FCFEFE}{\color[HTML]{000000} .02} & \cellcolor[HTML]{EFFAFA}{\color[HTML]{000000} .09} & \cellcolor[HTML]{F4FCFC}{\color[HTML]{000000} .06} & \cellcolor[HTML]{6ACAD1}{\color[HTML]{000000} .81} & \cellcolor[HTML]{FEFFFF}{\color[HTML]{000000} .01} & \cellcolor[HTML]{FEFFFF}{\color[HTML]{000000} .01} \\
    {\color[HTML]{000000} \textsc{disgust}} & \cellcolor[HTML]{FAFEFE}{\color[HTML]{000000} .03} & \cellcolor[HTML]{F3FBFC}{\color[HTML]{000000} .07} & \cellcolor[HTML]{E6F6F8}{\color[HTML]{000000} .14} & \cellcolor[HTML]{F4FCFC}{\color[HTML]{000000} .06} & \cellcolor[HTML]{7ED1D8}{\color[HTML]{000000} .70} & \cellcolor[HTML]{FEFFFF}{\color[HTML]{000000} .01} \\
    {\color[HTML]{000000} \textsc{fear}} & \cellcolor[HTML]{FCFEFE}{\color[HTML]{000000} .02} & \cellcolor[HTML]{E7F7F8}{\color[HTML]{000000} .13} & \cellcolor[HTML]{F1FAFB}{\color[HTML]{000000} .08} & \cellcolor[HTML]{F6FCFD}{\color[HTML]{000000} .05} & \cellcolor[HTML]{F8FDFD}{\color[HTML]{000000} .04} & \cellcolor[HTML]{82D3D9}{\color[HTML]{000000} .68} \\
    \hline
    {\color[HTML]{000000} \textsc{neutral}} & \cellcolor[HTML]{D3F0F2}{\color[HTML]{000000} .24} & \cellcolor[HTML]{B9E6EA}{\color[HTML]{000000} .38} & \cellcolor[HTML]{D7F1F3}{\color[HTML]{000000} .22} & \cellcolor[HTML]{F1FAFB}{\color[HTML]{000000} .08} & \cellcolor[HTML]{F8FDFD}{\color[HTML]{000000} .04} & \cellcolor[HTML]{FAFEFE}{\color[HTML]{000000} .03}\\
    \hline
    \end{tabular}

    \caption{.
    Distribution statistics between 
    speaker state ($u_{state}$) and 
    emotion \textit{speaker causes} ($e^{u\to*}$) in the competition training data (\textsc{train}\textsubscript{json});
    values in each row are normalized}
    \label{tab:train_analysis_se}
\end{table}

According to the Table~\ref{tab:train_analysis_dist},
most emotion was found to be caused by such utterances $u^{src}$ that are
the same as or mentioned before $u^{tgt}$ ($\delta \geq 0$).  
Therefore, 
given $\left<u^{src}, u^{tgt}\right>$
we denote its context
$X = \{u^1 \ldots u^k\}$
as a \textit{history} of the past $k-1$
utterances of $u^{tgt}$, where $u^{tgt}=u^k \in X$, $u^{src} \in X$.
\\
\textbf{Task definition:} Given an emotion-causing utterance pair within context
$\left< u^{src}, u^{tgt}, X\right>$ answer the emotion $e_{c}~\in~\labelsAllMath{}$ caused by $u^{src}$ towards $u^{tgt}$.

\section{Methodology}
\label{sec:methodology}

We propose a two-stage training mechanism
for performing
instruction-tuning
on
large language models (LLMs),
aimed at accurately inferring of the task answers.
Given triplet $\left<u^{src}, u^{tgt}, X\right>$
of emotion-cause pair 
$\left<u^{src}, u^{tgt}\right>$
in context 
$X$, the proposed mechanism aims at LLM instruction-tuning, in order to answer $e \in E'$ that refers to: 
\begin{enumerate}[label={\textsc{stage \arabic*}:},leftmargin=2.1cm]
    \item emotion state $u^{tgt}_{state}$;
    \item emotion cause by $u^{src}$ to $u^{tgt}$. 
\end{enumerate}
Therefore, for emotion-cause pairs extraction we use the \textsc{stage 2} towards the model tuned in \textsc{stage 1} to infer $e_{c}~\in~\labelsAllMath{}$ caused by $u^{src}$ towards $u^{tgt}$.

\begin{figure*}[!t]
 \centering
  \includegraphics[width=\textwidth]{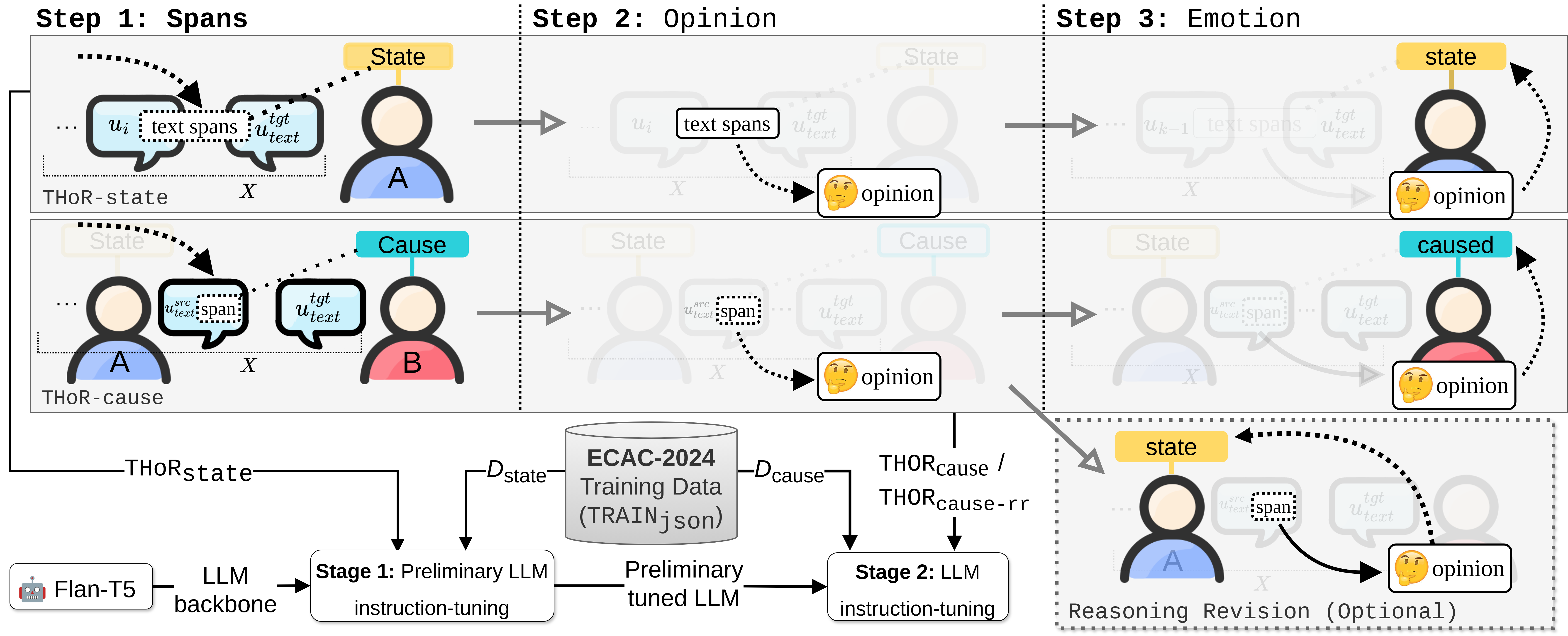}
  \caption{Two-stage LLM tuning methodology for inferring emotion caused by $u^{src}$ towards $u^{tgt}$ in context $X$ by adapting \thor{}~\cite{FeiAcl23THOR} 
 to reason and answer: (i) $u^{tgt}_{state}$ 
  (\thorState{}), and (ii) emotion caused by $u^{src}$ towards $u^{tgt}$ (\thorCause{}), optionally enhanced by Reasoning-Revision and by predicting $u^{src}_{state}$ (\thorCauseRR{}).}
  \label{fig:three-hop-methodology-adaptation}
\end{figure*}

Instead of directly asking LLM the final result at each stage, we exploit the Chain-of-Thought (CoT) concept in the form of the Three-hop Reasoning (\textsc{THoR}) framework~\cite{FeiAcl23THOR}. 
We believe that LLM can infer the span that conveys emotion and opinion about it before answering $e\in \labelsAllMath{}$.
Figure~\ref{fig:three-hop-methodology-adaptation} illustrates the proposed training methodology, empowered by the CoT prompting.
We refer to the instruction-tuning mechanisms of the
\textsc{stage~1} and 
\textsc{stage~2}
as
\thorState{} and
\thorCause{}
respectively.

\subsection{Chain-of-Thought Prompting}
\label{sec:cot-prompting}

We adopt the \thor{} framework~\cite{FeiAcl23THOR} in LLM fine-tuning with the prompt templates adapted for 
emotion-cause pair analysis in conversations.
We define the intermediate \textit{span} ($s$) and latent \textit{opinion} expression ($o$). 
With $C_{i}, i\in \overline{1..3}$ we denote the prompts that wrap the content in the input context.
The construction of stages is as follows.

\textbf{\thorState{}}
This is a \textsc{stage 1} of the proposed training methodology, aimed at 
preliminary LLM instruction-tuning. 
Given $\left<u^{tgt}, X\right>$, we apply the following three steps to infer $u^{tgt}_{state} = e'_{1} \in \labelsAllMath{}$: 
\begin{tcolorbox}[breakable,colframe=gray,boxrule=2pt,arc=3.4pt,boxsep=-1mm]
    \textbf{Step~1}:%
    {
    $s'_1$ =
    [$C_1(X)$, which text spans are possibly causes emotion on $u^{tgt}_{text}$?]
    }  
\end{tcolorbox}
\begin{tcolorbox}[breakable,colframe=gray,boxrule=2pt,arc=3.4pt,boxsep=-1mm]
    \textbf{Step~2}: $o'_{1}$ = 
    [$C_2(C_1, s'_{1})$. 
    Based on the common sense, what is the implicit opinion towards the mentioned text spans that causes emotion on $u^{tgt}_{text}$, and why?]
\end{tcolorbox}
\begin{tcolorbox}[breakable,colframe=gray,boxrule=2pt,arc=3.4pt,boxsep=-1mm]
    \textbf{Step~3}: $e'_{1}=$
    [$C_3(C_2, o'_{1})$. Based on such opinion, what is the emotion state of $u^{tgt}_{text}$?]
\end{tcolorbox}
where $s'_{1}$ could be interpret as 
$s'_{1} = argmax~p(s_{1}|X,u^{tgt}_{text})$,
latent opinion $o'_{1}$ as $o'_{1} = argmax~p(o_{1}|X,u^{tgt}_{text},s'_{1})$, and
the final answer $e'_{1}$ noted as: $e'_{1}= argmax~p(e_{1}|X,u^{tgt}_{text},s'_{1},o'_{1})$.

\textbf{\thorCause{}} 
This is a \textsc{stage 2} of the proposed methodology, based on emotions-cause pairs.
We use this stage for (i)~\textit{fine-tuning} and (ii)~task result \textit{inferring} purposes.
Given context $\left<u^{src}, u^{tgt}, X\right>$ 
we omit\footnote{To reduce the problem statement to the one for which \thor{} was originally designed~\cite{pontiki2016semeval}} $u^{tgt}\in X$ from the input parameters by referring to it as <<\textit{end of the conversation}>>.
We apply the following steps to infer 
$e_{2}' \in \labelsAllMath{}$ 
caused by $u^{src}$ to $u^{tgt}$:
\begin{tcolorbox}[breakable,colframe=gray,boxrule=2pt,arc=3.4pt,boxsep=-1mm]
    \textbf{Step 1}:~$s'_{2}$ = 
    [$C_1(X)$, which specific text span of $u^{src}_{text}$ is possibly causes emotion?]
\end{tcolorbox}
\begin{tcolorbox}[breakable,colframe=gray,boxrule=2pt,arc=3.4pt,boxsep=-1mm]
    \textbf{Step 2}:~$o'_{2}$ = 
    [$C_2(C_1, s_{2}')$. Based on the common sense, what is the implicit opinion towards the cause of mentioned text span of $u^{src}_{text}$, and why?]
\end{tcolorbox}
\begin{tcolorbox}[breakable,colframe=gray,boxrule=2pt,arc=3.4pt,boxsep=-1mm]
    \textbf{Step 3}:~$e'_{2}$ = 
    [$C_3(C_2, o'_{2})$. Based on such opinion, what is the emotion caused by {source} towards the last conversation utterance?]
\end{tcolorbox}
where $s'_{2}$ could be interpret as 
$s'_{2} = argmax~p(s'_{2}|X,u^{src}_{text})$,
opinion $o'$ could be interpret as 
$o'_{2} = argmax~p(o_{2}|X,u^{src}_{text},s'_{2})$, 
and the  final answer $e'_{2}$ noted as: 
$e'_{2}~=~argmax~p(e_{2}|X,u^{src}_{text},s'_{2},o'_{2})$.

\subsection{Reasoning Revision with Supervision}
\label{sec:reasoning-revision}

During the LLM instruction-tuning process with the \thor{}, it is possible to devise a reasoning path.
Technically, at each step of the chain we have all the necessary information to query our model with the final answer. 
With the following approach, we believe in a better model alignment on state-cause dependency (Table~\ref{tab:train_analysis_se}):
speakers are likely to cause an emotion, similar to their states\footnote{Except \textsc{neutral} speaker state (Table~\ref{tab:train_analysis_se})}.
To revise this knowledge, in this paper, we impute the following prompt to support our opinion $O$, obtained at the end of the \thorCause{} step~2 (Fig.~\ref{fig:three-hop-methodology-adaptation}):
\begin{tcolorbox}[breakable,colframe=gray,boxrule=2pt,arc=3.4pt,boxsep=-1mm]
\textbf{Step 3.1}: $u'^{src}_{state} = $[$C_3$($C_2$, $o'_{2}$), Based on such opinion, what is the emotion state of $u^{src}_{text}$?]
\end{tcolorbox}
Due to the definition of the task, 
we believe in the correctness of this knowledge within the emotion cause task. 
Once step 3.1 is embedded,
the result answer
$e'_{2} \in E'$ 
in
\thorCause{}
from the step~3
could be reinterpret as
$e'_{2}~=~argmax~p(e_{2}|X, u^{src}_{text}, s'_{2}, o'_{2}, u'^{src}_{state})$.
We refer to this setup as
\thorCauseRR{}.

\section{Datasets and Experiential Setup}
\label{sec:dataset}


We adopt textual resources provided by the competition organizers~\cite{wang-EtAl:2024:SemEval20244}:
training (\trainData{}) and
evaluation (\testData{}) data.
Within \trainData{}, for each conversation,
we rely on
(i)~speakers \textit{emotion states}, and
(ii)~\textit{emotion causes} annotation
to compose
the datasets \dState{} and \dCause{}, respectively. 
Each dataset represent a list of tuples $\dTuple{} = (u, X, \dLabels{})$, 
where $u$ is an utterance of the conversation context
$X~=~\{u^1 \ldots u^k\}$,  
and $\dLabels{}$ is a list of emotion labels, defined as:
\begin{itemize}
    \item $\dLabels{} = [u^{k}_{state}]$ in the case of \dState{} ($u^k_{state} \in \labelsAllMath{}$)
    \item $\dLabels{} = [u_{state}, e^{u}]$ in the case of \dCause{}, where $e^{u}$ is emotion expressed by $u$ towards $u^k$, or {\small \textsc{neutral}} otherwise ($e^{u} \in \labelsAllMath{}$)
\end{itemize}


\dState{} represent entries of all possible utterances in all conversations with their emotional states $u_{state} \in \labelsMath{}$.
For the particular utterance $u$,
we consider its context as $X_u = \{u': u_{id} - u'_{id} \leq k\}$.

\begin{table}[!t]
    \centering
    \small
    \begin{tabu}{l|cc|c}
    Source         & \multicolumn{2}{c|}{\textsc{train}\textsubscript{json}}  & \textsc{test}\textsubscript{json} \\
    \hline
    Part                & \texttt{train} &   \texttt{dev}  &  \texttt{test} \\
    \hline
    \dState{}   (total)                   & 12144 & 	1475 &  
    \\
    \hspace{3mm}\textsc{neutral}            & 5299  & 	630     &  \textcolor{gray}{.}   \\
    \hspace{3mm}\textsc{joy}                & 2047  & 	254     &  \textcolor{gray}{.} \\
    \hspace{3mm}\textsc{surprise}           & 1656  & 	184     &  \textcolor{gray}{.} \\
    \hspace{3mm}\textsc{anger}              & 1423  & 	192    &  \textcolor{gray}{.} \\
    \hspace{3mm}\textsc{sadness}            & 1011  & 	136     &  \textcolor{gray}{.} \\
    \hspace{3mm}\textsc{disgust}            & 372   & 	42     &  \textcolor{gray}{.} \\
    \hspace{3mm}\textsc{fear}               & 336   & 	37     &  \textcolor{gray}{.} \\
    \hline
    \dCause{} (total)                      & 30445 & 	3612 & 	15794                 \\
    \hspace{3mm}\textsc{neutral}             & 23750&  2765 	     &  15794                 \\
    \hspace{3mm}\textsc{joy}                 & 2111 &  279  	     &  \textcolor{gray}{--}   \\
    \hspace{3mm}\textsc{surprise}            & 1725 &  202  	     &  \textcolor{gray}{--}   \\
    \hspace{3mm}\textsc{anger}               & 1307  &  174  	     &  \textcolor{gray}{--}   \\
    \hspace{3mm}\textsc{sadness}             &932   &  120  	     &  \textcolor{gray}{--}   \\
    \hspace{3mm}\textsc{disgust}             & 387  &   47  	     &  \textcolor{gray}{--}   \\
    \hspace{3mm}\textsc{fear}                & 233   &   25  	     &  \textcolor{gray}{--}   \\
    \hline
    \end{tabu}
    \caption{Statistics of the composed datasets \dState{} and \dCause{} from the publicly available competition data, for the two training methodology stages respectively; statistics is listed for $k=3$.}
    \label{tab:datasets}
\end{table}

\dCause{} includes all possible pairs
$\left<u^{src}, u^{tgt}\right>$, 
where 
$u^{src}_{id} \leq u^{tgt}_{id}$, 
and 
$u^{tgt}_{id} - u^{src}_{id} \leq k$.
For the particular pair, we compose the related context ($X'$) as follows: $X'=\{u': u^{tgt}_{id} - u'_{id} \leq k\}$.
For each pair, we assign $e\in \labelsMath{}$ if the pair is present in conversation annotation and \textsc{\small neutral} otherwise.
We rely on the analysis in Table~\ref{tab:train_analysis_dist} to limit the number of pairs, as well as the size of the context.
We set $k~=~3$ to cover $95.8$\% emotion-cause pairs.
We also cover the case of emotions caused from within the same utterance 
($59.5$\%, see Table~\ref{tab:train_analysis_s}).
As for
emotions caused by the same speaker of other utterance, 
we assess that
excluding this type of pairs
($12.83\%$, according to Table~\ref{tab:train_analysis_s}), 
results in 
$\approx 23\%$ 
pairs 
reduction of \dCause{} and hence reduces training time.
Therefore, the result \dCause{} excludes pairs of this type in 
\texttt{train}, \texttt{dev} and \texttt{test}
parts.

Table~\ref{tab:datasets} lists the statistics of the composed resources.
We use the 9:1 proportion for
\trainData{}
to compose \texttt{train} and \texttt{dev}, respectively.
To represent $X\in \dTuple{}$, we concatenate its representation of utterances.
For each utterance $u~\in~X$, 
we use the following formatting template: <<{$u_{speaker}$ : $u_{text}$}>>.
To represent utterance $u \in \dTuple{}$, we refer to $u_{text}$.
For each $l \in L$ formatting, we utilize its lowercase text value.
The implementation details for the datasets preparation are publicly available.\footnote{\scriptsize \url{https://github.com/nicolay-r/SemEval2024-Task3}}
\\
\textbf{Setup.} We 
follow the publicly available framework setups~\cite{FeiAcl23THOR} and
adopt encoder-decoder style instructive Flan-T5\footnote{\scriptsize\url{https://huggingface.co/google/flan-t5-base}} 
as our backbone LLM for the proposed methodology.
We experiment with a 250M (base) version.
For evaluations on \texttt{dev}, 
we adopt the F1-measure for $\labelsAllMath{}$, denoted as \fdev{}.
The evaluation on \texttt{test} assessed with the set of $F1$-metrics,
provided by the competition organizers (details in Section~\ref{sec:experiments}).
We
consider the instruction-tuning of the Flan-T5 model with the following techniques:
conventional 
\textsc{prompt}, 
\thor{} (Section~\ref{sec:cot-prompting}), and 
\textsc{\thorCause{}} with reasoning revision (Section~\ref{sec:reasoning-revision}).
To conduct the experiment, we rent a server with a single
NVIDIA A100 GPU (40GB).
We set 
temperature 
$1.0$, 
learning rate 
$2\cdot 10^{-4}$,
optimizer AdamW~\cite{loshchilov2017decoupled}, 
\textsc{batch-size} of $32$.

For the \textsc{prompt} technique, we use the template
{<<$C_1(X)$. $I(u)$. Choose from $\labelsAllMath{}$>>},
where $I(u)$ corresponds to the instruction.
For \dCause{} we use $I(u)=$~{<<What emotion causes $u_{text}$ towards the last conversation utterance?>>}

\section{Experiments}
\label{sec:experiments}


\begin{figure}[!t]
    \centering
    \includegraphics[width=0.95\linewidth,trim={2mm 1mm 2mm 0mm},clip]{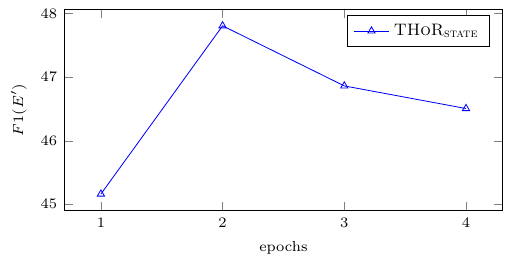}
    \caption{
    Result analysis 
    of the  
    preliminary fine-tuning of
    Flan-T5\textsubscript{base} 
    on 
    \dState{} 
    \texttt{dev} 
    using 
    \thorState{}
    technique per epoch by 
    \fdev{}
    }
    \label{fig:analysis-emotion-state}

    \bigskip 
    
    \centering
    \includegraphics[width=0.95\linewidth,trim={2mm 1mm 2mm 0mm},clip]{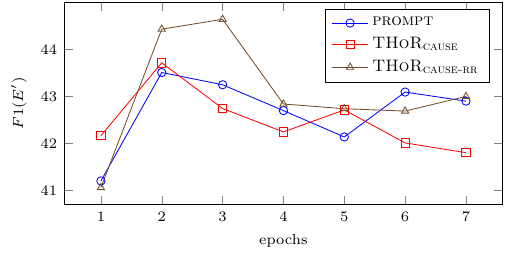}
    \caption{
    Flan-T5\textsubscript{base}\textdagger{}
    fine-tuning results comparison 
    by \fdev{}
    on \dCause{}~\texttt{dev} part per 
    each epoch 
    across fine-tuning techniques:
    \textsc{prompt}, 
    \thorCause{}, 
    and
    \thorCauseRR{}.
    }
    \label{fig:analysis-emotion-cause}
\end{figure}

\textbf{Stage 1}.
Figure~\ref{fig:analysis-emotion-state} illustrates the analysis of the 
$F1$ on \texttt{dev} part during the preliminary tuning
of Flan-T5\textsubscript{base}
on \dState{}.\footnote{We left the comparison with other pre-training techniques listed in~\ref{sec:dataset} out of scope of this paper due to alignment with the CoT concept in \textsc{stage 2}.}
We investigate the overfitting after $2$ epochs of training.
The best state, obtained at the end of the epoch~\#2 
with the \fdev{}$=47.81$ on the \dState{}-\texttt{dev} part,
has been selected. 
In further, we refer to this model as
Flan-T5\textsubscript{base}\textdagger{}.
\\
\textbf{Stage 2}
Figure~\ref{fig:analysis-emotion-cause} provides a comparative analysis of different fine-tuning techniques.
As at the pre-training stage, we investigate the ability to learn task emotion states 2-3 training epoch, followed by overfitting.
Switching from \textsc{prompt} to \thorCauseRR{} technique,
we investigate the improvement by $2.5$\% percent by \fdev{} on the \texttt{dev} part
of the \dState{} dataset. 
We refer to the best fine-tuned versions as Flan-T5\textsubscript{base}\textdaggerdbl{},
separately per each fine-tuning technique in Table~\ref{tab:results} (\texttt{dev} column).

The official evaluation includes the following $F1$ measures: 
(i)~weighted averaged $F1^{w}_{*}$ / non-weighted ($F1_{*}$), 
and 
(ii)~strict ($F1_{s}$) / not-strict ($F1_{p}$) towards predicted spans. 
To form the submissions for official evaluation, 
the following span corrections approaches were used:
(i)~punctuation terms\footnote{We use \texttt{string.punctuation}  preset in Python}
exclusion from utterance prefixes and suffixes (by default), and (ii)~algorithm-based (Section~\ref{sec:spans-correction}).
Table~\ref{tab:results} (\texttt{test} columns) illustrate the available results of 
T5\textsubscript{base}\textdaggerdbl{} in official evaluation.
\\
\textbf{Final submission} represents the results of Flan-T5\textsubscript{base}\textdaggerdbl{} 
(\textsc{THOR\textsubscript{cause-rr}} technique),
and application of algorithm-based spans correction.

\begin{table}[!t]
    \centering
    \setlength{\tabcolsep}{4pt}
    \begin{tabular}{lccccc}
    \hline
    Source & \texttt{\small dev} & \multicolumn{4}{c}{\small \texttt{test}}\\
    \hline
    Model & \fdev{} & $F1^{w}_{s}$ & $F1^{w}_{p}$& $F1_{s}$& $F1_{p}$\\
    \hline
    \multicolumn{6}{l}{\textbf{\textsc{prompt}}}\\
    FT5\textsubscript{base}\textdaggerdbl{} & 43.51 &   9.68  &  22.27   &  10.05  &  22.21   \\
    \multicolumn{6}{l}{\textbf{\textsc{THoR\textsubscript{cause}}}}\\
    FT5\textsubscript{base}\textdaggerdbl{} & 43.72 &  \textcolor{gray}{--} &  \textcolor{gray}{--} & \textcolor{gray}{--} &  \textcolor{gray}{--}\\
    \multicolumn{6}{l}{\textbf{\textsc{THoR\textsubscript{cause-rr}}}}\\
    FT5\textsubscript{base}\textdaggerdbl{} & 44.64 &  9.74 &  23.54 &  10.33  &  23.94 \\
    \hline
    \multicolumn{6}{l}{\textbf{\textsc{THoR\textsubscript{cause-rr}} + {\small \textit{Algorithm-based Spans Correction}}}}\\
    \rowcolor{light-gray} 
    FT5\textsubscript{base}\textdaggerdbl{} & 44.64 &  12.86   &  24.28   &  13.26   &   24.13  \\
    \hline
    \end{tabular}
    \caption{
        Evaluation results for 
        Flan-T5\textsubscript{base}\textdaggerdbl{}
        on
        \texttt{dev} and \texttt{test} parts 
        of the \dCause{} dataset; the results of the final submission are highlighted in gray}
    \label{tab:results}
\end{table}%
\begin{table}[!t]
    \centering
    \small

    \begin{tabular}{p{6cm}|r}
        Parameter           & Value \\
        \hline
        Conversations (total)                   & 2917 \\
        Emotion causes pairs in annotation      & 665  \\
        \hspace{2mm} Average per conversation   & 4.39 \\
        \hline
    \end{tabular}

    \caption{Quantitative statistics of the 
    automatically extracted emotion-cause pairs by 
    Flan-T5\textsubscript{base}\textdaggerdbl{}
    (\textsc{THOR\textsubscript{cause-rr}} technique)
    from the evaluation data (\textsc{test}\textsubscript{json})
    }
    \label{tab:final-submission-analysis-q}
    
    \bigskip
    
    \begin{tabular}{p{3cm}|rrrr}
    Parameter & \multicolumn{4}{l}{past} \\
    \hline
    $\delta = u^{tgt}_{id}-u^{src}_{id}$  & 0    & 1    & 2    & 3    \\
    \hline
    Causes count       & 1711 & 1012 & 148  & 46   \\
    \hspace{3mm}Average per $\delta$  & 2.57 & 1.52 & 0.22 & 0.07 \\
    \hspace{3mm}Covering (\%)      & 58.7 & 93.3 & 98.4 & 100.0  \\
    \hline
    \end{tabular}
    
    \caption{
    Statistic of distances in utterances ($\delta$) between source ($u^{src}$) and target ($u^{tgt}$) of emotion-cause pairs 
    for
    automatically extracted emotion-cause pairs by 
    Flan-T5\textsubscript{base}\textdaggerdbl{}
    (\textsc{THOR\textsubscript{cause-rr}} technique)
    from the evaluation data (\textsc{test}\textsubscript{json})
    }
    \label{tab:final-submission-analysis-dist}
    \bigskip
    
    \setlength{\tabcolsep}{4pt} 
    
    \begin{tabular}{l|llllll}
        $u_{state} \backslash e^{u\to*}$         & {\color[HTML]{000000} \textsc{joy}}                 & {\color[HTML]{000000} \textsc{sur}}                & {\color[HTML]{000000} \textsc{ang}}                    & {\color[HTML]{000000} \textsc{sad}}                     & {\color[HTML]{000000} \textsc{dis}}                     & {\color[HTML]{000000} \textsc{fea}}                        \\
        \hline
        {\color[HTML]{000000} \textsc{joy}}      & \cellcolor[HTML]{5FC6CE}{\color[HTML]{000000} .87} & \cellcolor[HTML]{F1FAFB}{\color[HTML]{000000} .08} & \cellcolor[HTML]{FCFEFE}{\color[HTML]{000000} .02} & \cellcolor[HTML]{FEFFFF}{\color[HTML]{000000} .01} & \cellcolor[HTML]{FEFFFF}{\color[HTML]{000000} .01} & \cellcolor[HTML]{FFFFFF}{\color[HTML]{000000} .00} \\
        {\color[HTML]{000000} \textsc{surprise}} & \cellcolor[HTML]{EFFAFA}{\color[HTML]{000000} .09} & \cellcolor[HTML]{75CED5}{\color[HTML]{000000} .75} & \cellcolor[HTML]{F4FCFC}{\color[HTML]{000000} .06} & \cellcolor[HTML]{F6FCFD}{\color[HTML]{000000} .05} & \cellcolor[HTML]{FAFEFE}{\color[HTML]{000000} .03} & \cellcolor[HTML]{FEFFFF}{\color[HTML]{000000} .01} \\
        {\color[HTML]{000000} \textsc{anger}}    & \cellcolor[HTML]{F6FCFD}{\color[HTML]{000000} .05} & \cellcolor[HTML]{E6F6F8}{\color[HTML]{000000} .14} & \cellcolor[HTML]{82D3D9}{\color[HTML]{000000} .68} & \cellcolor[HTML]{F1FAFB}{\color[HTML]{000000} .08} & \cellcolor[HTML]{FAFEFE}{\color[HTML]{000000} .03} & \cellcolor[HTML]{FEFFFF}{\color[HTML]{000000} .01} \\
        {\color[HTML]{000000} \textsc{sadness}}  & \cellcolor[HTML]{F4FCFC}{\color[HTML]{000000} .06} & \cellcolor[HTML]{EBF8F9}{\color[HTML]{000000} .11} & \cellcolor[HTML]{FAFEFE}{\color[HTML]{000000} .03} & \cellcolor[HTML]{73CDD4}{\color[HTML]{000000} .76} & \cellcolor[HTML]{FCFEFE}{\color[HTML]{000000} .02} & \cellcolor[HTML]{FCFEFE}{\color[HTML]{000000} .02} \\
        {\color[HTML]{000000} \textsc{disgust}}  & \cellcolor[HTML]{F3FBFC}{\color[HTML]{000000} .07} & \cellcolor[HTML]{EBF8F9}{\color[HTML]{000000} .11} & \cellcolor[HTML]{F3FBFC}{\color[HTML]{000000} .07} & \cellcolor[HTML]{F6FCFD}{\color[HTML]{000000} .05} & \cellcolor[HTML]{82D3D9}{\color[HTML]{000000} .68} & \cellcolor[HTML]{FEFFFF}{\color[HTML]{000000} .01} \\
        {\color[HTML]{000000} \textsc{fear}}     & \cellcolor[HTML]{FFFFFF}{\color[HTML]{000000} .00} & \cellcolor[HTML]{E4F6F7}{\color[HTML]{000000} .15} & \cellcolor[HTML]{EFFAFA}{\color[HTML]{000000} .09} & \cellcolor[HTML]{FCFEFE}{\color[HTML]{000000} .02} & \cellcolor[HTML]{FFFFFF}{\color[HTML]{000000} .00} & \cellcolor[HTML]{77CFD5}{\color[HTML]{000000} .74} \\
        \hline
        {\color[HTML]{000000} \textsc{neutral}}  & \cellcolor[HTML]{BDE8EB}{\color[HTML]{000000} .36} & \cellcolor[HTML]{B5E5E9}{\color[HTML]{000000} .40} & \cellcolor[HTML]{F3FBFC}{\color[HTML]{000000} .07} & \cellcolor[HTML]{E9F8F9}{\color[HTML]{000000} .12} & \cellcolor[HTML]{FAFEFE}{\color[HTML]{000000} .03} & \cellcolor[HTML]{FCFEFE}{\color[HTML]{000000} .02}\\ 
        \hline
    \end{tabular}
    
    \caption{
    Distribution statistics between 
    speaker state ($u_{state}$) and 
    emotion \textit{speaker causes} ($e^{u\to*}$) 
    for
    automatically extracted emotion-cause pairs by 
    Flan-T5\textsubscript{base}\textdaggerdbl{}
    (\textsc{THOR\textsubscript{cause-rr}} technique)
    from the evaluation data (\textsc{test}\textsubscript{json});
    values in each row are normalized  
    }
    \label{tab:final-submission-analysis-caused-to}
    \vspace{3mm}
    
    \setlength{\tabcolsep}{4pt} 
    
    \begin{tabular}{l|llllll}
     $u_{state} \backslash e^{*\to u}$          & {\color[HTML]{000000} \textsc{joy}}                         & {\color[HTML]{000000} \textsc{sur}}                    & {\color[HTML]{000000} \textsc{ang}}                       & {\color[HTML]{000000} \textsc{sad}}                     & {\color[HTML]{000000} \textsc{dis}}                     & {\color[HTML]{000000} \textsc{fea}}                        \\
    \hline
    {\color[HTML]{000000} \textsc{joy}}      & \cellcolor[HTML]{FFD86B}{\color[HTML]{000000} .97} & \cellcolor[HTML]{FFFFFE}{\color[HTML]{000000} .01} & \cellcolor[HTML]{FFFFFE}{\color[HTML]{000000} .01} & \cellcolor[HTML]{FFFFFF}{\color[HTML]{000000} .00} & \cellcolor[HTML]{FFFFFE}{\color[HTML]{000000} .01} & \cellcolor[HTML]{FFFFFF}{\color[HTML]{000000} .00} \\
    {\color[HTML]{000000} \textsc{surprise}} & \cellcolor[HTML]{FFFEF9}{\color[HTML]{000000} .04} & \cellcolor[HTML]{FFDB77}{\color[HTML]{000000} .89} & \cellcolor[HTML]{FFFEF9}{\color[HTML]{000000} .04} & \cellcolor[HTML]{FFFFFE}{\color[HTML]{000000} .01} & \cellcolor[HTML]{FFFFFE}{\color[HTML]{000000} .01} & \cellcolor[HTML]{FFFFFE}{\color[HTML]{000000} .01} \\
    {\color[HTML]{000000} \textsc{anger}}    & \cellcolor[HTML]{FFFEF9}{\color[HTML]{000000} .04} & \cellcolor[HTML]{FFFDF8}{\color[HTML]{000000} .05} & \cellcolor[HTML]{FFDD81}{\color[HTML]{000000} .83} & \cellcolor[HTML]{FFFDF8}{\color[HTML]{000000} .05} & \cellcolor[HTML]{FFFFFC}{\color[HTML]{000000} .02} & \cellcolor[HTML]{FFFFFE}{\color[HTML]{000000} .01} \\
    {\color[HTML]{000000} \textsc{sadness}}  & \cellcolor[HTML]{FFFFFC}{\color[HTML]{000000} .02} & \cellcolor[HTML]{FFFFFC}{\color[HTML]{000000} .02} & \cellcolor[HTML]{FFFEFB}{\color[HTML]{000000} .03} & \cellcolor[HTML]{FFDB77}{\color[HTML]{000000} .89} & \cellcolor[HTML]{FFFFFC}{\color[HTML]{000000} .02} & \cellcolor[HTML]{FFFFFE}{\color[HTML]{000000} .01} \\
    {\color[HTML]{000000} \textsc{disgust}}  & \cellcolor[HTML]{FFFFFC}{\color[HTML]{000000} .02} & \cellcolor[HTML]{FFFEF9}{\color[HTML]{000000} .04} & \cellcolor[HTML]{FFFDF8}{\color[HTML]{000000} .05} & \cellcolor[HTML]{FFFDF5}{\color[HTML]{000000} .07} & \cellcolor[HTML]{FFDE84}{\color[HTML]{000000} .81} & \cellcolor[HTML]{FFFFFE}{\color[HTML]{000000} .01} \\
    {\color[HTML]{000000} \textsc{fear}}     & \cellcolor[HTML]{FFFFFF}{\color[HTML]{000000} .00} & \cellcolor[HTML]{FFFDF6}{\color[HTML]{000000} .06} & \cellcolor[HTML]{FFFDF5}{\color[HTML]{000000} .07} & \cellcolor[HTML]{FFFEF9}{\color[HTML]{000000} .04} & \cellcolor[HTML]{FFFEFB}{\color[HTML]{000000} .03} & \cellcolor[HTML]{FFDF85}{\color[HTML]{000000} .80} \\
    \hline
    {\color[HTML]{000000} \textsc{neutral}}  & \cellcolor[HTML]{FFD666}{\color[HTML]{000000} .60} & \cellcolor[HTML]{FFF7DE}{\color[HTML]{000000} .13} & \cellcolor[HTML]{FFFDF8}{\color[HTML]{000000} .03} & \cellcolor[HTML]{FFF5D7}{\color[HTML]{000000} .16} & \cellcolor[HTML]{FFFCF3}{\color[HTML]{000000} .05} & \cellcolor[HTML]{FFFEFA}{\color[HTML]{000000} .02}  \\
    \hline
    \end{tabular}
    
    \caption{
    Distribution statistics between 
    speaker state ($u_{state}$) 
    and 
    emotion \textit{caused on them} ($e^{*\to u}$),
    for
    automatically extracted emotion-cause pairs by 
    Flan-T5\textsubscript{base}\textdaggerdbl{}
    (\textsc{THOR\textsubscript{cause-rr}} technique)
    from the evaluation data (\textsc{test}\textsubscript{json});
    values in each row are normalized  
    }
    \label{tab:final-submission-analysis-caused-on-them}
\end{table}%

\subsection{Algorithm-based Spans Correction}
\label{sec:spans-correction}
Our methodology (Section~\ref{sec:methodology}) is limited on utterance level emotion cause prediction.\footnote{
Technically it is possible to obtain spans (Section~\ref{sec:methodology}), 
however we could not investigate the practical valuty of the 
\thorCause{}-based Flan-T5\textsubscript{base}\textdaggerdbl{} 
responses from step~\#1.}
We believe it is reflected in the relatively low results of $F1_{s}$ 
on the \texttt{test} dataset (see Table~\ref{tab:results}).
Therefore, we analyze \trainData{}
and adopt a placeholder solution, 
aimed at enhancing the results by $F1_{s}$.
\begin{algorithm}[!t]
\newcommand{\Break}{\State \textbf{break} }
\caption{\small Emotion-cause prefixes correction for $u_{text}$}
\label{alg:rule-based-method}
\small
\begin{algorithmic}
\State $updated \gets True$
\State $V_{p}^{'} \gets \text{sorted}~V_{p}~\text{by decreased entry lengths in words}$
\While{$u_{text} \neq \varnothing$ \textbf{or} $updated$}
    \State $updated \gets False$
    \State $u'_{text} \gets u_{text}$ \Comment{Modified version of $u_{text}$}
    \For{$v_{p} \in V_{p}^{'}$} 
        \If{$u_{text}$ ends with $v_{p}$}
            \State $u'_{text} \gets \text{part of}~u_{text}~\text{before}~v_{p}$
            \State $updated \gets True$
            \Break
     \EndIf \EndFor \EndWhile
\end{algorithmic}
\end{algorithm}

We apply a \textit{rule-based approach} 
based on differences between the original utterance texts and their span annotations in the training data.
Using
\trainData{}, we compose 
\textit{prefix-} ($V_p$) 
and 
\textit{suffix-} ($V_s$) 
vocabularies. 
For vocabulary entries, we select those that satisfy all of the following criteria:
(i)~the length of entry does not exceed $5$ words,
(ii)~entry starts 
(in the case of $V_s$), or ends
(in the case of $V_p$)
with the punctuation sign\textsuperscript{7}.

For each utterance text ($u_{text}$) that causes emotion, 
we compose an updated $u_{text}'$ by applying:
(1)~correction of $u_{text}$ prefixes with $V_p$, followed by 
(2)~correction of suffixes from $V_s$ for the results from (1).
We alter $u_{text}'$ in the case of
$u_{text}'=\emptyset$.
The algorithm~\ref{alg:rule-based-method}
illustrates an implementation for the prefixes correction with~$V_p$.\footnote{
Implementation is publicly available in 
\url{https://github.com/nicolay-r/SemEval2024-Task3}}

\subsection{Final Submission Analysis}
We report the following 
emotion-cause pairs $\left<u^{src}, u^{tgt}\right>$
analysis results
for the
Flan-T5\textsubscript{base}\textdaggerdbl{} 
(\textsc{THOR\textsubscript{cause-rr}} technique, final submission):
\begin{enumerate}
    \item Quantitative statistics of the extracted emotion-cause pairs
    (Table~\ref{tab:final-submission-analysis-q});
    \item Distance statistics (in utterances) between 
    $u^{src}$ and $u^{tgt}$ (Table~\ref{tab:final-submission-analysis-dist});
    \item Distribution statistics between speaker state ($u_{state}$) and the emotion \textit{speaker causes} ($e^{u\to*}$)
    (Table~\ref{tab:final-submission-analysis-caused-to});
    \item Distribution statistics between speaker state ($u_{state}$) and emotion \textit{caused on them} ($e^{*\to u}$) (Table~\ref{tab:final-submission-analysis-caused-on-them}).
\end{enumerate}

According to the results in Table~\ref{tab:final-submission-analysis-caused-to},
we observe that 
the correlation between the state of the speaker $u$ utterance
($u_{state}$)
and the emotion it causes ($e^{u\to *}$) 
\textbf{is similar to} the related statistics on the competition training data (Table~\ref{tab:train_analysis_se}).
We also investigate the alignment of the speaker states
($u_{state}$)
with the emotion caused on them ($e^{* \to u}$) and the precision of the result varies between 80-97\% (Table~\ref{tab:final-submission-analysis-caused-on-them}).
The known source of misalignment is the case when emotion\footnote{\textsc{\small joy} especially, as the most frequently appearing class.}
$e^{*\to u} \in \labelsMath{}$
caused on $u$ with $u_{state} = \text{\small \textsc{neutral}}$
(bottom row, Table~\ref{tab:final-submission-analysis-caused-on-them}).

\section{Conclusion}

In this paper, we present a Chain-of-Thought (CoT) methodology aimed at
fine-tuning LLM for emotion state and cause extraction. 
We consider the problem of 
\textit{emotion cause analysis in conversations}
as a context-based problem with the mentioned utterance that causes emotion towards the last utterance in context.
We devise our CoT for emotion causes and propose a reasoning revision methodology aimed at imputing the speaker emotion to support the decision on caused emotion.
Our CoT represent a Three-hop Reasoning approach priory known as \thor{}.
We apply this approach to fine-tune LLM and predict: 
(i)~emotion state of the mentioned utterance, and (ii)~emotion caused by mentioned utterance towards the last utterance in context. 
We experiment with the Flan-T5\textsubscript{base} (250M) model fine-tuning 
using resources provided by task organizers.
The application of CoT with reasoning revision allows us
to improve the results by $2.5$\% (F1-measure) compared to prompt-based tuning. 
In further work, we expect to contribute with the:
(i)~analysis of larger models, and 
(ii)~enhanced reasoning revision techniques, mentioned in the
final submission analysis.

\bibliography{main} 

\begin{thebibliography}{6}
\expandafter\ifx\csname natexlab\endcsname\relax\def\natexlab#1{#1}\fi

\bibitem[{Hao et~al.(2023)Hao, Bobo, Qian, Lidong, Fei, and Tat-Seng}]{FeiAcl23THOR}
Fei Hao, Li~Bobo, Liu Qian, Bing Lidong, Li~Fei, and Chua Tat-Seng. 2023.
\newblock Reasoning implicit sentiment with chain-of-thought prompting.
\newblock In \emph{Proceedings of the Annual Meeting of the Association for Computational Linguistics}, pages 1171--1182.

\bibitem[{Loshchilov and Hutter(2017)}]{loshchilov2017decoupled}
Ilya Loshchilov and Frank Hutter. 2017.
\newblock Decoupled weight decay regularization.
\newblock \emph{arXiv preprint arXiv:1711.05101}.

\bibitem[{Pontiki et~al.(2016)Pontiki, Galanis, Papageorgiou, Androutsopoulos, Manandhar, AL-Smadi, Al-Ayyoub, Zhao, Qin, De~Clercq et~al.}]{pontiki2016semeval}
Maria Pontiki, Dimitris Galanis, Haris Papageorgiou, Ion Androutsopoulos, Suresh Manandhar, Mohammed AL-Smadi, Mahmoud Al-Ayyoub, Yanyan Zhao, Bing Qin, Orph{\'e}e De~Clercq, et~al. 2016.
\newblock Semeval-2016 task 5: Aspect based sentiment analysis.
\newblock In \emph{ProWorkshop on Semantic Evaluation (SemEval-2016)}, pages 19--30. Association for Computational Linguistics.

\bibitem[{Wang et~al.(2023)Wang, Ding, Xia, Li, and Yu}]{wang2023multimodal}
Fanfan Wang, Zixiang Ding, Rui Xia, Zhaoyu Li, and Jianfei Yu. 2023.
\newblock Multimodal emotion-cause pair extraction in conversations.
\newblock \emph{IEEE Transactions on Affective Computing}, 14(3):1832--1844.

\bibitem[{Wang et~al.(2024)Wang, Ma, Xia, Yu, and Cambria}]{wang-EtAl:2024:SemEval20244}
Fanfan Wang, Heqing Ma, Rui Xia, Jianfei Yu, and Erik Cambria. 2024.
\newblock \href {https://aclanthology.org/2024.semeval2024-1.273} {Semeval-2024 task 3: Multimodal emotion cause analysis in conversations}.
\newblock In \emph{Proceedings of the 18th International Workshop on Semantic Evaluation (SemEval-2024)}, pages 2022--2033, Mexico City, Mexico. Association for Computational Linguistics.

\bibitem[{Xia and Ding(2019)}]{xia2019emotion}
Rui Xia and Zixiang Ding. 2019.
\newblock Emotion-cause pair extraction: A new task to emotion analysis in texts.
\newblock In \emph{Proceedings of the 57th Annual Meeting of the Association for Computational Linguistics}, pages 1003--1012.

\end{thebibliography}

\end{document}